\documentclass{article}
\usepackage{amssymb}
\usepackage{amsmath}
\usepackage[ruled,vlined]{algorithm2e}
\usepackage{enumitem}
\usepackage{booktabs}
\usepackage{makecell} 
\usepackage[table]{xcolor}
\usepackage{xcolor}
\usepackage{wrapfig}
\usepackage{tabularx}
\usepackage{tikz}
\usetikzlibrary{shapes, arrows.meta, positioning}
\definecolor{tablecolor}{RGB}{255,224,179}
\definecolor{mygreen}{RGB}{0, 128, 0} 
\definecolor{myred}{RGB}{255, 0, 0}   

\newcommand{\singledd}[1]{$\footnotesize #1$}
\newcommand{\singleddbf}[1]{\cellcolor{tablecolor}$\footnotesize \mathbf{#1}$}
\usepackage[preprint]{corl_2025} 

\title{DemoSpeedup: Accelerating Visuomotor Policies via Entropy-Guided Demonstration Acceleration}

\author{
Lingxiao Guo$^{1}$, \quad
Zhengrong Xue$^{2,1,3}$, \quad
Zijing Xu$^{4}$, \quad
Huazhe Xu$^{2,1,3}$ \\
\vspace{0.1cm} 
$^{1}$ Shanghai Qi Zhi Institute, $^{2}$ Tsinghua University IIIS, $^{3}$Shanghai AI Lab, \\$^{4}$University of Electronic Science and Technology of China\\
\texttt{guolingxiao3@gmail.com,
huazhe\_xu@mail.tsinghua.edu.cn}
}

\begin{document}
\maketitle


\vspace{-0.4cm}
\begin{abstract}
Imitation learning has shown great promise in robotic manipulation, but the policy's execution is often unsatisfactorily slow due to commonly tardy demonstrations collected by human operators.
In this work, we present \textit{DemoSpeedup}, a self-supervised method to accelerate visuomotor policy execution via entropy-guided demonstration acceleration.
\textit{DemoSpeedup} starts from training an arbitrary generative policy (e.g., ACT or Diffusion Policy) on normal-speed demonstrations, which serves as a per-frame action entropy estimator.
The key insight is that frames with lower action entropy estimates call for more consistent policy behaviors, which often indicate the demands for higher-precision operations. In contrast, frames with higher entropy estimates correspond to more casual sections, and therefore can be more safely accelerated. 
Thus, we segment the original demonstrations according to the estimated entropy, and accelerate them by down-sampling at rates that increase with the entropy values.
Trained with the speedup demonstrations, the resulting policies execute up to $3$ times faster while maintaining the task completion performance.
Interestingly, these policies could even achieve higher success rates than those trained with normal-speed demonstrations, due to the benefits of reduced decision-making horizons. Project Page: \url{https://demospeedup.github.io/}.


\end{abstract}

\keywords{Imitation Learning, Manipulation, Demonstration Acceleration} 


\begin{figure}[h]
\vspace{-5pt}
  \centering
  \makebox[\linewidth]{\hspace*{-0.05\linewidth}
  \includegraphics[width=1.1\linewidth]{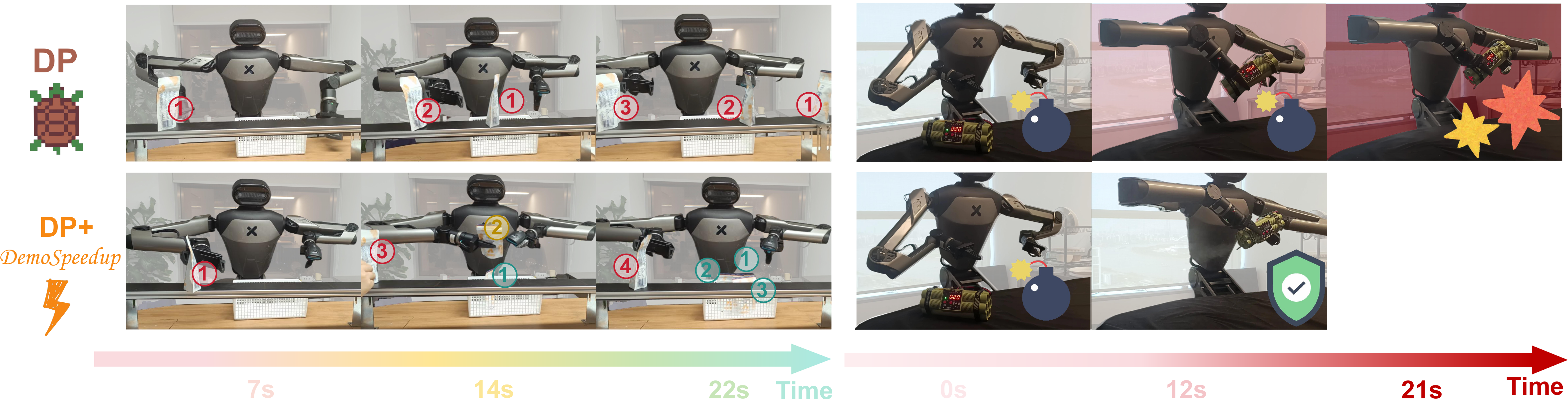}}
\caption{Manipulation speed is crucial for improving the productivity and ensuring the success of time-sensitive tasks. \textbf{\textit{DemoSpeedup}} enables boosting the execution speed of visuomotor policy from slow demonstrations. Left: Original policy fails to track the speed of the production line and scan the products, while {\textit{DemoSpeedup}} succeeds. Right: {\textit{DemoSpeedup}} succeed to depose the time bomb toy before the end of countdown but the original policy fails.}
  \label{fig:teaser}
\vspace{1pt}
\end{figure}
\section{Introduction}
	
Imitation learning has achieved remarkable success for robot manipulation tasks from a perspective of task completion rates~\citep{chi2023diffusion,black2410pi0,zhao2024aloha,ze20243d}, but visuomotor policies are often less satisfactory in terms of time efficiency. For visually pleasing fluency, it has become a common practice in the policy learning community that the presented video demos are accelerated by $2\times\sim10\times$~\cite{chi2023diffusion,black2410pi0,xue2025demogen}. However, low efficiency might be a concern for some time-sensitive settings in the real world, e.g., caring for babies and the elderly or manufacturing on an assembly line.

Recently, some test-time policy acceleration techniques have been developed to improve execution speed by naively down-sampling the action chunk to be executed at test time~\citep{Helix, xie2024subconscious}. However, test-time acceleration often leads to a notable performance drop when the acceleration rate is relatively high (e.g., by $2\times$) due to the apparent distributional shift induced by speedup during deployment.

In this work, we assume that tardy demonstrations collected by human operators are the primary cause for the slow execution of behavior cloning policies. Empirically, we observed that even skillful data collectors with over $100$ hours of experience struggle to teleoperate the robot arms as fluently as using their own hands. In both VR~\citep{cheng2024open,ding2024bunny,ze2024generalizable} and kinematics-teaching~\citep{fu2024mobile,wu2024gello,yang2024ace,jiang2025behavior} teleoperation, the lack of full-directional, non-occluded view of the 3D scene, as well as the absence of tactile proprioception, are the major obstacles to achieving faster motions. Besides, morphological heterogeneity between humans and robots, combined with equipment latency, further increases the difficulty of teleoperation and slows down the data collection speed. 

In response to the tardiness of human-collected demonstrations, we introduce \textit{DemoSpeedup}, designed to boost policies' execution efficiency without sacrificing their task completion performance.
Rather than naively downsample the demonstrated trajectories by a constant rate, \textit{DemoSpeedup} preserves the high-precision sections (e.g., picking up objects) and only accelerates in the low-precision sections 
(e.g., approaching objects in free air) 
to promote the task completion rate after acceleration.

The core of \textit{DemoSpeedup} is an entropy-guided precision measurement mechanism, which allows the adaptive adjustment of the acceleration rate while avoiding the need for additional human annotations or hand-designed, task-specific priors. Our key observation is that human operators tend to have multiple casual yet reasonable choices in low-precision sections, and follow more consistent behaviors in high-precision sections to ensure successful manipulation. Therefore, \textit{action entropy}, which reflects the randomness of the actions, could be an implicit indicator of the precision required.
However, the discretized actions recorded in the demonstration trajectory are very sparsely located in the multi-dimensional continuous action space, which is the major obstacle to calculating the action entropy offline from the human-collected demonstration dataset.

In \textit{DemoSpeedup}, we propose to overcome the obstacle by estimating action entropy from a self-supervised proxy policy, which can be an arbitrary generative behavior learning policy trained on the non-accelerated source dataset. The proxy policy is not responsible for action prediction, but is only used to help distill the action entropy information embedded in the source dataset. A clustering-based scheme is designed to process the proxy-inferred per-frame entropy into trajectory segmentation with precision labeling, ready for piecewise varying-speed acceleration. Finally, the accelerated dataset facilitates the training of an accelerated behavior cloning policy, which is the end product of the \textit{DemoSpeedup} pipeline used for action prediction during deployment.

Empirically, we validate the effectiveness of \textit{DemoSpeedup} by instantiating it with two popular generative behavior cloning policies: Action Chunking with Transformers (ACT)~\citep{zhao2023learning} and Diffusion Policy (DP)~\citep{chi2023diffusion}. We conduct extensive experiments on a diverse range of tasks in the simulation and the real world. The results demonstrate \textit{DemoSpeedup} strikes a $1.7\times\sim3\times$ speedup in time efficiency, while obtaining task completion success rates on par with or sometimes even higher than the same policy trained on the non-accelerated dataset.

\section{Related Work}
    \textbf{Learning from human demonstrations. }
    Learning from human demonstrations has emerged as a widely adopted approach in robotic manipulation. Generative policies trained by imitation learning, such as ACT~\citep{zhao2023learning} and Diffusion Policy~\citep{chi2023diffusion} can strike a performance that matches the demonstrations. The generalization of imitation learning~\citep{intelligence2025pi_, qu2025spatialvla,liu2025hybridvla} has been boosted over the years. However, execution speed of current imitation learning paradigms is still restricted by the slow demonstrations. The problem still exists in large datasets~\citep{bu2025agibot,wu2024robomind,khazatsky2024droid,o2024open} collected by teleoperation. While VLAs trained with those large datasets~\citep{bu2025agibot,bjorck2025gr00t,kim2024openvla,li2024cogact} exhibits strong generalization, numerous slow demonstrations within the data cause the learned policies to be much slower than normal human speed. This hinders the real-world application of robots in daily life.

    \textbf{Data curation for manipulation. }
    Several prior works curate data aiming to improve the success rate of the trained policy. \citep{chen2025curating} train and cross-validate a classifier to discern successful policy roll-outs from unsuccessful ones and use the classifier to filter heterogeneous demonstration datasets. \citep{shi2023waypoint, belkhale2023hydra} down-sample the demonstrations with either geometric or human-labeled metrics, showing that by reducing episode length, the compounding error could drop~\citep{belkhale2023data} and the success rate can increase. But they relies on close-loop control and make the manipulation time longer. Overall, all these methods focus on improving success rates. In contrast, the objective of curating data in this work is to boost manipulation speed. At the same time, the episode length reduces, resulting in a potential effect to improve the performance. 
     
\section{Method}
\begin{figure}[t]

\vspace{-0.1in}
    \centering
    \includegraphics[width=\textwidth]{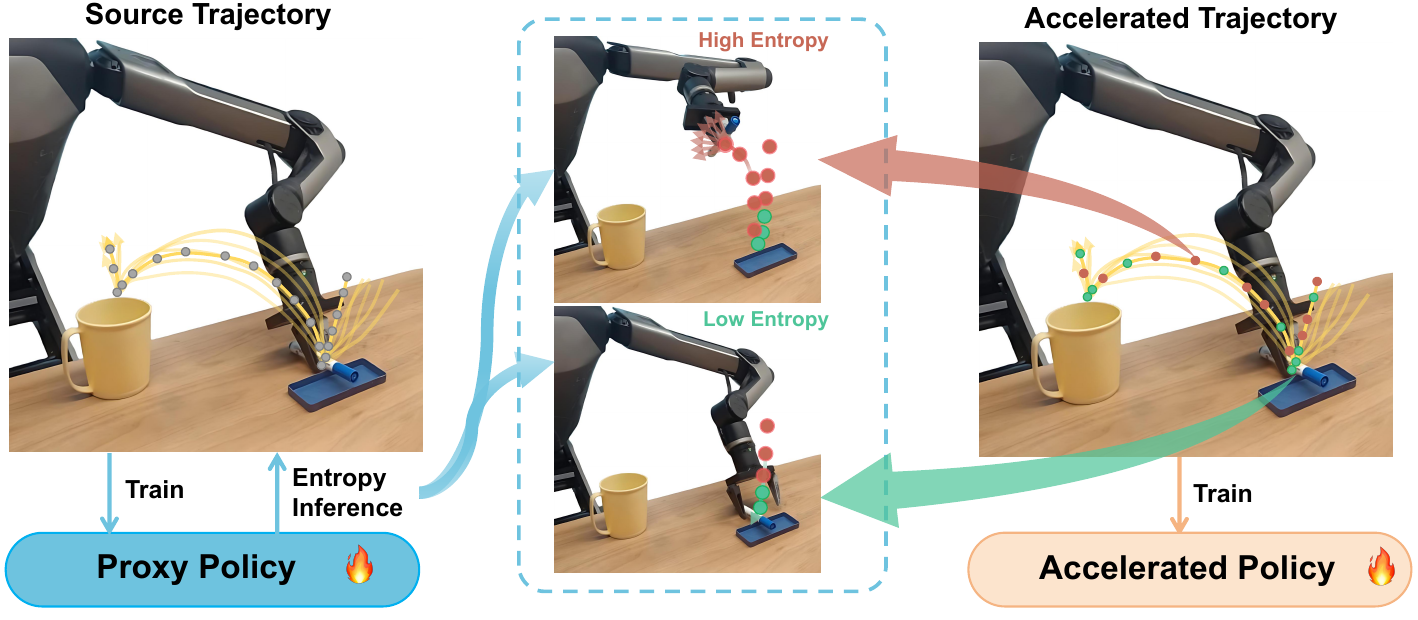}
    \caption{\textbf{\textit{DemoSpeedup}} utilizes a generative policy trained from original demonstrations to estimate conditional action entropy.  Actions with high entropy \textcolor{myred}{(red points)} are down-sampled at a higher rate while actions with low entropy \textcolor{mygreen}{(green points)}  are down-sampled at a lower rate.} 
    \label{fig:pipeline}
    \vspace{-1em}
\end{figure}
In this section, we present \textit{DemoSpeedup}, an entropy-guided approach to accelerate demonstrations to speed up policy execution. As shown in Figure \ref{fig:pipeline}, \textit{DemoSpeedup} first utilizes a proxy policy trained on source data to estimate the action entropy in a nonparametric way. Then it leverages a density-based cluster method to segment trajectories into high-precision and low-precision sets, followed by piecewise down-sampling at rates according to the precision level. Finally, some training and deployment details is introduced to guarantee the performance of accelerated policies.
\subsection{Action Entropy Estimation}
We leverage action entropy to reflect the precision level required for an action in human demonstrations.
To estimate the conditional action entropy of demonstration frames, \textit{DemoSpeedup} starts from training a proxy policy on the source dataset of original-speed demonstrations. 
We represent the behavior cloning policy as $\pi_\theta(A_t|o_t)$, where $A_t = \{a_t[t],a_t[t+1],...,a_t[t+K-1]\}$ is the action chunk~\citep{lai2022action} and $K$ is the chunk length. 
For entropy calculation, we sample $N$ action chunk samples $\{a^i_t[t],...,a^i_t[t+K-1]\}_{i=1}^N$ conditioned on the current observation $o_t$.  Then, we perform Gaussian kernel density estimation \citep{heer2021fast} to obtain the probability density distribution of the actions conditioned on the current observation: \begin{equation}\hat{p}(a_t|o_t) = \frac{1}{N Kh} \sum_{j=t-K+1}^{t} \sum_{i=1}^{N} \frac{1}{\sqrt{2 \pi}} \exp\left( -\frac{(a_t - a^i_j[t])^2}{2 h^2} \right)\label{equation1}\end{equation}
where $h$ is the bandwidth. Then we estimate the conditional action entropy at $o_t$ by:
\begin{equation}\hat{H}(a_t|o_t) = - \sum_{j=t-K+1}^{t} \sum_{i=1}^{N} \hat{p}(a^i_j[t]|o_t) \log \hat{p}(a^i_j[t]|o_t)\label{equation2}\end{equation}
We perform the per-frame operation along all timesteps for all the trajectories in the dataset. We instantiate the proxy policy with either Action Chunking with Transformers (ACT)~\citep{zhao2023learning} or Diffusion Policy (DP)~\citep{chi2023diffusion}. For ACT, action samples are obtained by sampling different latent variables in the CVAE prior distribution, i.e, $x \sim \mathcal{N}(0, 1)$. For DP, action samples are generated by sampling multiple noise sequences given the observation. 

\subsection{Entropy-Guided Demonstration Acceleration}
The estimated entropy paves the way for subsequent steps to identify the precision level of different segments and accelerate them at different rates. We develop a cluster-based approach to determine the precision level and leverage entropy for demonstration speedup. 

\textbf{Entropy preprocessing.} As the teleoperation data can be very noisy and have harmful impact on clustering, we first utilize Isolation Forest~\citep{xu2017improved} to detect the abnormal entropy values in one trajectory, after which the outliers are substituted by the adjacent normal values. Then, the entropy of each frame $\hat{H}(a_t|o_t)$ is first concatenated with its time index $t$ to preserve the temporal property. All these obtained entropy points in one episode are normalized for clustering preparation.

\textbf{Clustering for precision labeling.} We adopt a density-based clustering method to divide those entropy points into fine-grained and coarse-grained areas. 
Specifically, we adopt hierarchical density-based clustering (Hdbscan)~\citep{mcinnes2017hdbscan} to cluster those entropy points. Those high-entropy points are labeled to outliers, while low-entropy areas are labeled to clusters. To further exclude clustering noise, we simply filter all the obtained clusters by preserving clusters in which the mean entropy values are lower than zero. All the time indices in the preserved clusters are labeled as set $P$, i.e., precision set; and the rest are identified as set $C$, i.e., casualness set.

\textbf{Replicate-before-downsample strategy.} After getting precision labels, now it's possible to speed up the temporal segments at different rates by down-sampling. However, naively down-sampling the whole trajectory $\{(o_t, a_t)\}_{t=1}^{T}$ will significantly reduce the visited state diversity in the dataset, causing a severe waste of the demonstrations and empirically leading to a serious performance drop. 
To avoid the potential information loss caused by acceleration, we develop a simple replicate-before-downsample strategy, which retains all the observation frames that appear in the source dataset.
More specifically, at an acceleration rate of $N\times$, the target chunk is replicated into $N$ copies. The $i$-th copy is down-sampled by $N\times$ with a starting offset of $i$ frames. Instead of skipping the intermediate frames, our strategy essentially splits the chunk into $N$ accelerated sub-chunks, thus retaining the same diversity of the visited states as in the source demonstrations.

\textbf{Geometrical consistency.} Since action chunking has crucial impact on imitation learning performance, it is necessary to determine the chunk length of the accelerated policy trained on speedup demonstrations. We opt to keep the geometrical distance traveled by an action chunk roughly the same as the original policy. This ensures that the accelerated policy only needs to fit much less action labels than the original policy for the same segments in the demonstrations, which benefits for converging and reducing compounding error. 

\textbf{Controller requirements.} During data collection and deployment, acceleration requires high-precision controller of the robot. Apparently, if the controller is inaccurate, the control dynamics could differ a lot between the original speed and speedup actions, which also leads to a performance drop. During evaluation, we find some robot gripper controllers fail to track high speed and cause failures. We simply increase the gripper gain to solve this problem.
\section{Experiments}

\begin{figure}
\vspace{-5pt}
  \centering
  \includegraphics[width=1.0\linewidth]{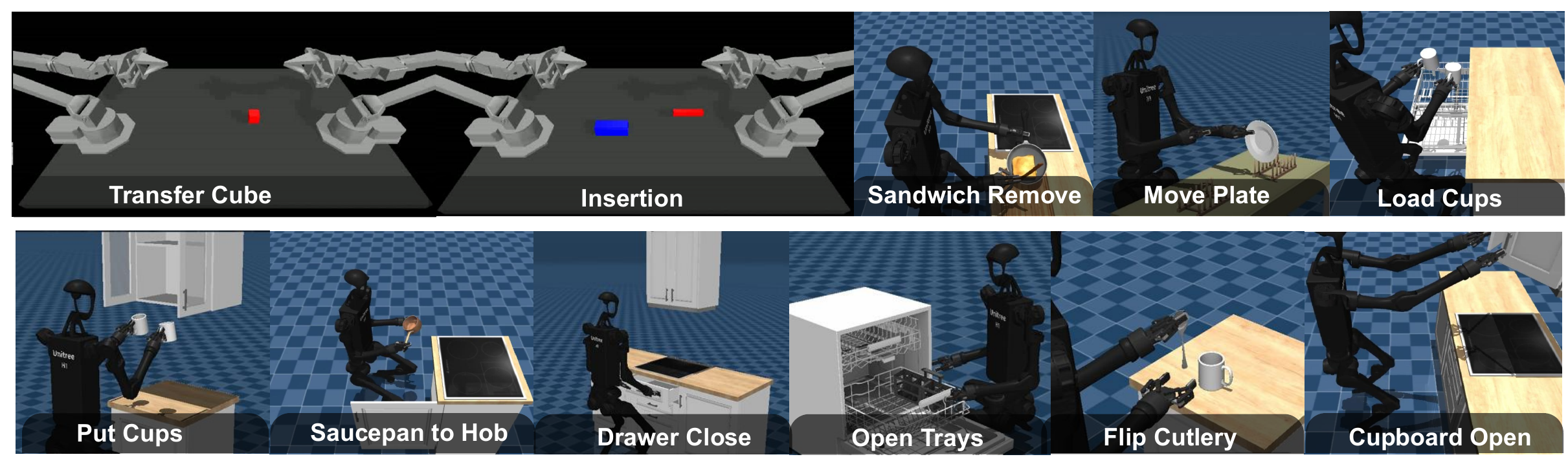}
    \vspace{-15pt}
\caption{\textbf{Simulation tasks. }The environments are from Aloha and Bigym, featuring bimanual and mobile manipulation from human-collected datasets.} 
  \label{fig:sim tasks}
  \vspace{-5pt}
\end{figure}
\begin{table}[tbp]
\centering
\footnotesize
\setlength{\tabcolsep}{2.7pt}
\begin{tabularx}{\textwidth}{@{}l *{12}{>{\centering\arraybackslash}X}@{}} 
\toprule
\textbf{Method} 
& \multicolumn{2}{c}{\textbf{\scriptsize Transfer Cube}} 
& \multicolumn{2}{c}{\textbf{\scriptsize Insertion}} 
& \multicolumn{2}{c}{\textbf{\scriptsize Sandwich Remove}} 
& \multicolumn{2}{c}{\textbf{\scriptsize Move Plate}} 
& \multicolumn{2}{c}{\textbf{\scriptsize Load Cups}} 
& \multicolumn{2}{c}{\textbf{\scriptsize Put Cups}} \\
\cmidrule(lr){2-3} \cmidrule(lr){4-5} \cmidrule(lr){6-7} \cmidrule(lr){8-9} \cmidrule(lr){10-11} \cmidrule(lr){12-13}
\addlinespace[-8pt]
& \makecell{\scriptsize success \\[-2.5pt] \scriptsize rate\scriptsize ($\uparrow$)} 
& \makecell{\scriptsize episode \\[-2.5pt] \scriptsize len\scriptsize ($\downarrow$)} 
& \makecell{\scriptsize success \\[-2.5pt] \scriptsize rate\scriptsize ($\uparrow$)} 
& \makecell{\scriptsize episode \\[-2.5pt] \scriptsize len\scriptsize ($\downarrow$)} 
& \makecell{\scriptsize success \\[-2.5pt] \scriptsize rate\scriptsize ($\uparrow$)} 
& \makecell{\scriptsize episode \\[-2.5pt] \scriptsize len\scriptsize ($\downarrow$)} 
& \makecell{\scriptsize success \\[-2.5pt] \scriptsize rate\scriptsize ($\uparrow$)} 
& \makecell{\scriptsize episode \\[-2.5pt] \scriptsize len\scriptsize ($\downarrow$)} 
& \makecell{\scriptsize success \\[-2.5pt] \scriptsize rate\scriptsize ($\uparrow$)} 
& \makecell{\scriptsize episode \\[-2.5pt] \scriptsize len\scriptsize ($\downarrow$)} 
& \makecell{\scriptsize success \\[-2.5pt] \scriptsize rate\scriptsize ($\uparrow$)} 
& \makecell{\scriptsize success \\[-2.5pt] \scriptsize rate\scriptsize ($\uparrow$)}  \\
\addlinespace[-8pt]
\midrule
ACT      & \singledd{72\%} & \singledd{291} & \singledd{21\%} & \singledd{452} & \singledd{53\%} & \singledd{368} & \cellcolor{yellow!50}\singleddbf{54\%} & \singledd{157} & \singleddbf{61\%} & \singledd{319} & \singledd{61\%} & \singledd{288} \\
ACT-2x   & \singledd{70\%} & \singledd{162} & \singledd{13\%} & \singledd{238} & \singledd{46\%} & \singledd{193} & \singledd{46\%} & \singledd{119} & \singledd{50\%} & \singledd{195} & \singledd{54\%} & \singledd{141} \\
ACT+\textit{\scriptsize DemoSpeedup} & \singleddbf{81\%} & \singleddbf{121} & \singleddbf{30\%} & \singleddbf{151} & \singleddbf{77\%} & \singleddbf{156} & \singledd{53\%} & \singleddbf{91} & \singledd{59\%} & \singleddbf{176} & \singleddbf{62\%} & \singleddbf{132} \\
\midrule
DP       & \singledd{66\%} & \singledd{281} & \singledd{16\%} & \singledd{431} & \singledd{52\%} & \singledd{352} & \singleddbf{52\%} & \singledd{170} & \singledd{15\%} & \singledd{419} & \singledd{12\%} & \singledd{386} \\
DP-2x    & \singledd{61\%} & \singledd{146} & \singledd{12\%} & \singledd{245} & \singledd{51\%} & \singledd{247} & \singledd{41\%} & \singledd{125} & \singledd{11\%} & \singledd{177} & \singledd{7\%} & \singledd{243} \\
DP+\textit{\scriptsize DemoSpeedup} & \singleddbf{74\%} & \singleddbf{107} & \singleddbf{29\%} & \singleddbf{218} & \singleddbf{54\%} & \singleddbf{217} & \singledd{49\%} & \singleddbf{113} & \singleddbf{38\%} & \singleddbf{171} & \singleddbf{21\%} & \singleddbf{205} \\
\bottomrule

\toprule
\textbf{Method} 
& \multicolumn{2}{c}{\textbf{\scriptsize Saucepan to Hob}} 
& \multicolumn{2}{c}{\textbf{\scriptsize Drawers Close}} 
& \multicolumn{2}{c}{\textbf{\scriptsize Open Trays}} 
& \multicolumn{2}{c}{\textbf{\scriptsize Flip Cutlery}} 
& \multicolumn{2}{c}{\textbf{\scriptsize Cupboard Open}} 
& \multicolumn{2}{c}{\textbf{\scriptsize \quad\quad Averaged \quad\quad}} \\
\cmidrule(lr){2-3} \cmidrule(lr){4-5} \cmidrule(lr){6-7} \cmidrule(lr){8-9} \cmidrule(lr){10-11} \cmidrule(lr){12-13}
\addlinespace[-8pt]
& \makecell{\scriptsize success \\[-2.5pt] \scriptsize rate\scriptsize ($\uparrow$)} 
& \makecell{\scriptsize episode \\[-2.5pt] \scriptsize len\scriptsize ($\downarrow$)} 
& \makecell{\scriptsize success \\[-2.5pt] \scriptsize rate\scriptsize ($\uparrow$)} 
& \makecell{\scriptsize episode \\[-2.5pt] \scriptsize len\scriptsize ($\downarrow$)} 
& \makecell{\scriptsize success \\[-2.5pt] \scriptsize rate\scriptsize ($\uparrow$)} 
& \makecell{\scriptsize episode \\[-2.5pt] \scriptsize len\scriptsize ($\downarrow$)} 
& \makecell{\scriptsize success \\[-2.5pt] \scriptsize rate\scriptsize ($\uparrow$)} 
& \makecell{\scriptsize episode \\[-2.5pt] \scriptsize len\scriptsize ($\downarrow$)} 
& \makecell{\scriptsize success \\[-2.5pt] \scriptsize rate\scriptsize ($\uparrow$)} 
& \makecell{\scriptsize episode \\[-2.5pt] \scriptsize len\scriptsize ($\downarrow$)} 
& \makecell{\scriptsize success \\[-2.5pt] \scriptsize rate\scriptsize ($\uparrow$)} 
& \makecell{\scriptsize speedup \\[-2.5pt] \scriptsize($\uparrow$)} \\
\addlinespace[-8pt]
\midrule
ACT      & \singledd{86\%} & \singledd{383} & \singleddbf{100\%} & \singledd{119} & \singleddbf{100\%} & \singledd{244} & \singleddbf{63\%} & \singledd{193} & \singleddbf{100\%} & \singledd{146} & \singledd{77\%} & \singledd{1.0\times} \\
ACT-2x   & \singledd{81\%} & \singledd{224} & \singledd{87\%} & \singledd{84} & \singledd{93\%} & \singledd{149} & \singledd{49\%} & \singleddbf{121} & \singledd{96\%} & \singledd{103} & \singledd{69\%} & \singledd{1.7\times} \\
ACT+\textit{\scriptsize DemoSpeedup} & \singleddbf{92\%} & \singleddbf{163} & \singleddbf{100\%} & \singleddbf{63} & \singleddbf{100\%} & \singleddbf{105} & \singledd{62\%} & \singledd{141} & \singleddbf{100\%} & \singleddbf{81} & \singleddbf{82\%} & \singleddbf{2.1\times} \\
\midrule
DP       & \singleddbf{79\%} & \singledd{324} & \singleddbf{96\%} & \singledd{114} & \singledd{94\%} & \singledd{245} & \singleddbf{22\%} & \singledd{175} & \singleddbf{100\%} & \singledd{181} & \singledd{55\%} & \singledd{1.0\times} \\
DP-2x    & \singledd{41\%} & \singledd{242} & \singledd{81\%} & 65 & \singledd{86\%} & \singledd{157} & \singledd{18\%} & \singledd{127} & \singledd{94\%} & \singledd{161} & \singledd{45\%} & \singledd{1.6\times} \\
DP+\textit{\scriptsize DemoSpeedup} & \singleddbf{79\%} & \singleddbf{169} & \singledd{89\%} & \singleddbf{59} & \singleddbf{96\%} & \singleddbf{138} & \singledd{17\%} & \singleddbf{98} & \singleddbf{100\%} & \singleddbf{103} & \singleddbf{59\%} & \singleddbf{1.9\times} \\
\bottomrule
\label{sim results}
\end{tabularx}
\caption{\textbf{Simulation Results.} \textit{DemoSpeedup} achieves remarkable speedup effects while maintaining comparable success rate across different robot platforms and tasks.}
\end{table}

\subsection{Simulation Experiments}
\textbf{Compared Methods. } We compare the accelerated policies trained with \textit{DemoSpeedup}-accelerated datasets against the same ACT or DP policies trained with the original-speed demonstrations. Additionally, we compare \textit{DemoSpeedup} to more straightforward test-time acceleration~\citep{Helix} that naively downsamples the action chunk during evaluation by $2\times$, which we call ACT-${2\times}$ and DP-${2\times}$.

\textbf{Tasks. } We consider a total of $11$ tasks selected from Aloha~\citep{zhao2023learning} and BiGym~\citep{chernyadev2024bigym}, shown in Fig~\ref{fig:sim tasks}. 
\textit{For Aloha}, we focus on the tasks with relatively high precision requirements. We select Transfer Cube and Insert, with $50$ human demonstrations provided for each task. 
\textit{For BiGym}, we focus on mobile manipulation and tasks with longer horizons. We improve ACT and DP to have better performance on mobile manipulation tasks by 1) transforming the base action space into position control;  2) replacing the Resnet~\citep{he2016deep} with Multi-view Vision Transformer~\citep{seo2023multi} in DP to enhance multi-view fusion ability. For fair evaluation, we replay the demonstrations provided in the benchmark and filter out the failed ones~\citep{abbeel2coarse}. Tasks with extremely low success rate ($<10\%$) are excluded. A total of 9 BiGym tasks are selected, with different numbers of demonstrations provided in the benchmark and control frequencies ranging from $20\mathrm{Hz}$ to $50\mathrm{Hz}$.


\textbf{Metrics. }For evaluation, we report the task completion \textbf{success rate} and the averaged \textbf{episode length} for a successful policy rollout to measure time efficiency. 
\textit{For Aloha}, we perform $50$ episode rollouts using the checkpoint with minimal validation loss~\citep{shi2023waypoint,zhao2023learning}. 
\textit{For BiGym}, we report the maximum success rate and corresponding average episode length among $50$ evaluations throughout the training. All the results are averaged across $3$ seeds.

\textbf{Results.}
The quantitative results are presented in Table~\ref{sim results}.
Compared to ACT or DP trained on original-speed demonstrations, the same policies trained with \textit{DemoSpeedup}-accelerated datasets achieve the shortest time to complete the tasks while maintaining comparable performance.
Overall, \textit{DemoSpeedup} achieves an average speedup of approximately $2\times$ across different task setups and algorithms, with a maximum speedup of $3\times$.
On the other hand, while test-time downsampling shortens the completion time to a certain extent, it causes an average performance drop of over $8\%$. This reveals the advantage of demonstration acceleration over test-time acceleration.

\begin{figure}
  \centering
  \includegraphics[width=1.0\linewidth]{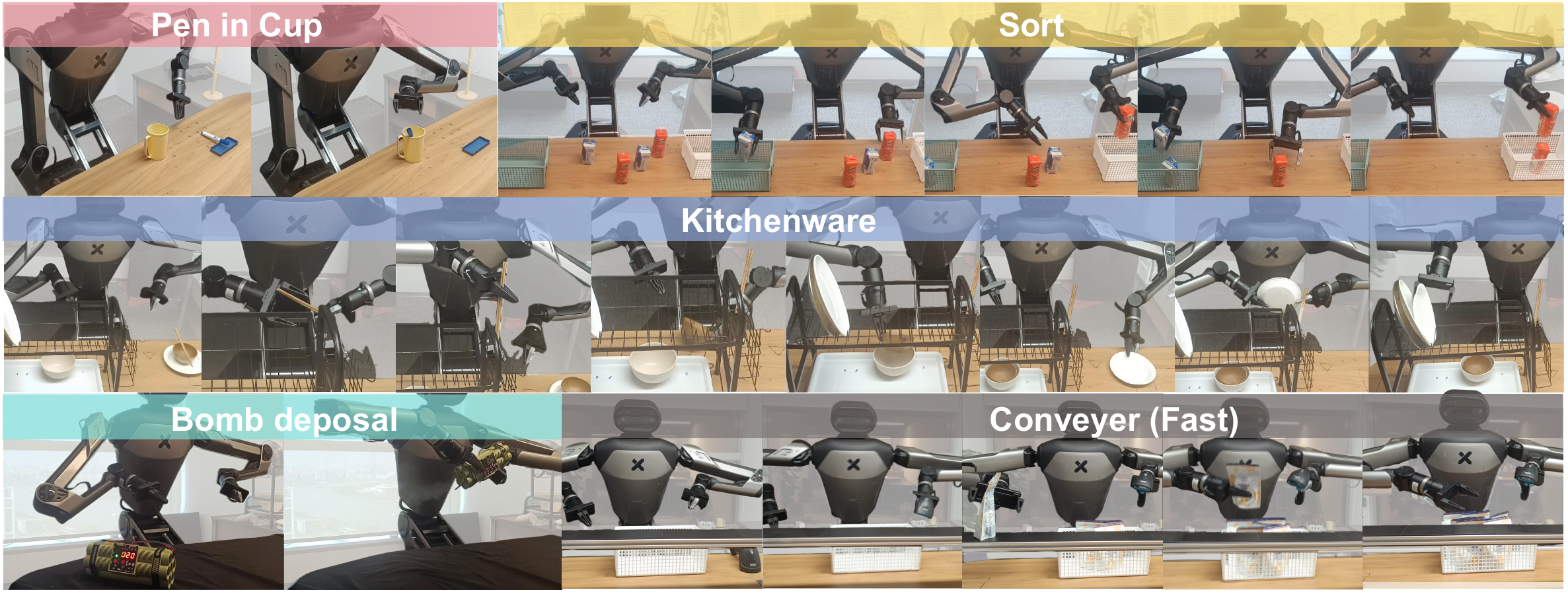}
\caption{\textbf{Real-world Setup.} We consider five real-world challenging tasks. \textit{Sort}, \textit{Kitchenware} emphasize long-horizon manipulation that require multiple skills. \textit{Bomb Deposal} requires precise manipulation. \textit{Conveyer} and its variation \textit{Conveyer Fast} is sensitive to manipulation speed.}
  \label{fig:real setup}
\vspace{-10pt}
\end{figure}
\subsection{Real-World Experiments}
\textbf{Tasks.} We design $5$ tasks and a variation on Galaxea R1, a bimanual humanoid platform. The tasks emphasize either long horizon or time sensitivity, as illustrated in Figure \ref{fig:real setup}. 
\begin{itemize}[leftmargin=20pt]
    \item \textit{Pen in Cup}: The robot needs to pick up a pen and place it inside of the cup. 
    \item \textit{Sort}: The robot is required to put all white yoghurt bottles into the green basket and all red ones into the white box. 
    \item \textit{Kitchenware}: The robot needs to grasp the chopsticks, bowl, and plate sequentially with its left arm, transfer them to the right arm, and then place them at the designated location. This is a long-horizon task requiring multiple skills like transferring and insertion.
    \item \textit{Bomb Deposal}: The robot needs to grasp the bomb toy, move it to its chest, and then precisely collaborate two arms to detach the battery wire. 
    \item \textit{Conveyer}: The robot is required to pick up the scanner, grasp the moving bag on the conveyor belt, scan the bag with the scanner, and then place the bag into the basket. Bags are continuously placed onto the conveyer belt by human.
    \item \textit{Conveyer Fast}: We evaluate the same checkpoints as in \textit{Conveyer} on a $2\times$ faster conveyer. It aims to simulate the situation where we want the robot more productive than the collected data.
\end{itemize}
For each task, the RGB visual observations are recorded through a Zed2 Camera mounted on the robot's head, and $100$ demonstrations are collected using the GalaxeaVR suite~\citep{GalaxeaVR}. The object configurations are randomized both in data collection and evaluation.

\textbf{Metrics.} We conduct $\sim30$ evaluation trials for each task, reporting the number of successful trials and the average time cost. 
The time cost is recorded by a stopwatch, which starts timing when the robot leaves its default joint positions and stops timing when the robot completes the task and returns to the default joint positions. 

\begin{table}[t]
\centering
\setlength{\tabcolsep}{3.pt} 
\small
\begin{tabularx}{\textwidth}{@{}l *{12}{>{\centering\arraybackslash}X}@{}} 
\toprule
\textbf{Method} 

& \multicolumn{2}{c}{\textbf{Pen in Cup}} 
& \multicolumn{2}{c}{\textbf{\quad\quad Sort\quad\quad\quad}} 
& \multicolumn{2}{c}{\textbf{Bomb Deposal}} 
& \multicolumn{2}{c}{\textbf{Kitchenware}} 
& \multicolumn{2}{c}{\textbf{Conveyer}}  
& \multicolumn{2}{c}{\textbf{Conveyer Fast}} \\ 
\cmidrule(lr){2-3} \cmidrule(lr){4-5} \cmidrule(lr){6-7} \cmidrule(lr){8-9} \cmidrule(lr){10-11} \cmidrule(lr){12-13} 
\addlinespace[-8pt]
\scriptsize
& \makecell{  \scriptsize  \quad success\\[-2.5pt] \scriptsize  rate ($\tiny\uparrow$)} 
& \makecell{  \scriptsize  \quad cost\\[-2.5pt]  \scriptsize \quad  time ($\tiny\downarrow$)} 
& \makecell{  \scriptsize  \quad success\\[-2.5pt] \scriptsize  rate ($\tiny\uparrow$)} 
& \makecell{  \scriptsize  \quad cost\\[-2.5pt]  \scriptsize \quad time ($\tiny\downarrow$)} 
& \makecell{ \scriptsize   \quad success\\[-2.5pt]  \scriptsize rate ($\tiny\uparrow$)} 
& \makecell{ \scriptsize  \quad  cost\\[-2.5pt]  \scriptsize \quad time ($\tiny\downarrow$)} 
& \makecell{ \scriptsize   \quad success\\[-2.5pt]  \scriptsize rate ($\tiny\uparrow$)} 
& \makecell{ \scriptsize  \quad  cost\\[-2.5pt] \scriptsize  \quad time ($\tiny\downarrow$)} 
& \makecell{  \scriptsize \quad  success\\[-2.5pt] \scriptsize  rate ($\tiny\uparrow$)} 
& \makecell{  \scriptsize  \quad cost\\[-2.5pt]  \scriptsize \quad time ($\tiny\downarrow$)} 
& \makecell{  \scriptsize  \quad success\\[-2.5pt]  \scriptsize rate ($\tiny\uparrow$)} 
& \makecell{  \scriptsize  \quad cost\\[-2.5pt] \scriptsize \quad time ($\tiny\downarrow$)} \\ 
\addlinespace[-6pt]
\midrule
ACT & \singledd{16/30} & \singledd{19.45s} & \singledd{29/40} & \singledd{56.78s} & \singleddbf{7/27} & \singledd{42.13s} & \singledd{6/33} & \singledd{66.32s} & \singledd{18/30} & \singledd{13.14s} & \singledd{2/30} & \singledd{12.68s} \\ 
ACT+\textit{\small Ours} & \singleddbf{24/30} & \singleddbf{8.28s} & \singleddbf{31/40} & \singleddbf{20.38s} & \singledd{6/27} & \singleddbf{26.31s} & \singleddbf{7/33} & \singleddbf{\hspace{-1.5pt} 27.26s} & \singleddbf{21/30} & \singleddbf{6.57s} & \singleddbf{16/30} & \singleddbf{6.28s} \\
\midrule
DP & \singledd{15/30} & \singledd{15.69s} & \singledd{32/40} & \singledd{39.29s} & \singledd{6/27} & \singledd{35.69s} & \singleddbf{19/33} & \singledd{61.12s} & \singleddbf{28/30} & \singledd{13.39s} & \singledd{7/30} & \singledd{12.96s} \\
DP+\textit{\small Ours} & \singleddbf{23/30} & \singleddbf{7.52s} & \singleddbf{\hspace{-1pt} 38/40} & \singleddbf{\hspace{-1pt}18.32s} & \singleddbf{11/27} & \singleddbf{19.18s} & \singledd{17/33} & \singleddbf{\hspace{-1pt}39.23s} & \singledd{25/30} & \singleddbf{6.24s} & \singleddbf{27/30} & \singleddbf{6.03s} \\
\bottomrule
\label{real results}
\end{tabularx}
\caption{\textbf{Real-World Results. }The results demonstrate the efficiency of \textit{DemoSpeedup} in accelerating the speed of visuomotor policies and the potential to improve the success rate.}
\end{table}
\textbf{Efficiency in boosting the speed of policy. }As shown in Table \ref{real results},  \textit{DemoSpeedup} achieves the lowest cost time among different tasks while maintaining the performance.  For tasks that require much accuracy such as Bomb Deposal and Kitchenware, \textit{DemoSpeedup} achieves at least 160\% speedup. For tasks that are not demanding on precision, \textit{DemoSpeedup} achieves a even higher speedup, such as 278\% for ACT and 214\% for DP in the Sort task. Besides, we notice that \textit{DemoSpeedup} obtains a higher speedup on ACT than DP. This is partly due to the DP inference delay. The sudden pause caused by the delay between faster movements can make the motions of DP a little more jittery, leading to a slight decrease in acceleration outcome.

\textbf{Potential for improving the success rate. }Interestingly, \textit{DemoSpeedup} could even boost the success rate in some tasks. We argue this is partially because \textit{DemoSpeedup} reduces the decision horizon, thereby reducing the compounding error in imitation learning~\citep{belkhale2023data}. Another reason is that when training policy with demonstrations, the change per timestep decreases proportionally as the speed is lower. Thus the marginal information of the action at each timestep is reduced, making it challenging for the policy training to converge and fit complex datasets~\citep{pertsch2025fast}. For example, in Pen in Cup tasks, the test positions of the cup are covered by the training data, but policies trained on original demonstrations are more likely to miss the correct position of the cup than those trained on speedup demonstrations. In addition, we observe that due to the real-world and Aloha demonstrations being slower than those in Bigym, the performance gains from \textit{DemoSpeedup} are more pronounced in the former two. This indicates that there is some correlation between the quality and speed of data, and \textit{DemoSpeedup} helps for fitting the dataset.
\begin{figure}
  \centering
  \includegraphics[width=1.0\linewidth]{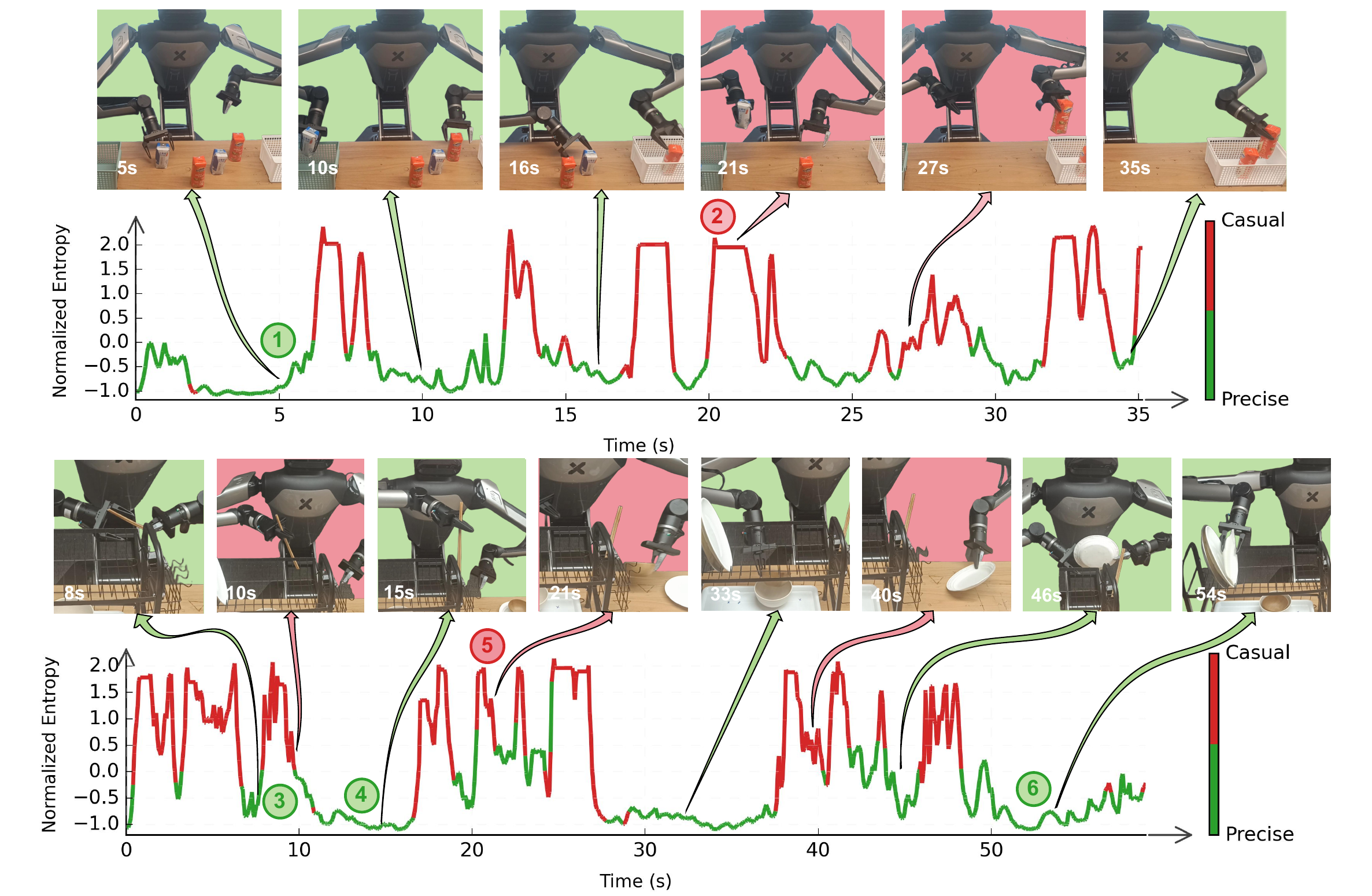}
\caption{\textbf{Entropy Visualization. }We showcase snapshots from the replayed demonstration and the corresponding normalized entropy curve of \textit{Sort}(upper row) and \textit{Kitchenware}(lower row). The green of the curve and the background stands for segmented precision set while red represents casualness set. The estimated entropy effectively captures both delicate skills and causal movements.} 
  \label{fig:entropy vis}
\end{figure}
\subsection{Ablations}
\begin{wraptable}[8]{r}{0.45\textwidth}  
    \centering
    \vspace{-12pt}
    \begin{tabular}{@{} l c c @{}}
        \toprule[0.4mm]
        Ablation & ACT & DP \\
        \midrule[0.2mm]
        \textit{DemoSpeedup} & \singleddbf{56\%} & \singleddbf{52\%} \\
        w/o. RBD strategy & \singledd{29}\% & \singledd{26\%} \\
        w/o. geometrical consistency  & \singledd{31\%} & \singledd{34\%} \\
        w/o. high precision ctrl  & \singledd{53\%} & \singledd{41\%} \\        
        \bottomrule
    \end{tabular}
    \caption{\textbf{Success rates on ablations.}}
    \label{table:ablation}
\end{wraptable}

We select two tasks from Aloha and utilize ACT and DP to conduct ablation studies in more details. Three designs are ablated: naively down-sample the whole trajectory instead of the replicate-before-downsample(RBD) strategy; adopt the same action chunk length instead of geometrically consistent chunk length; gripper without high-precision control. We report the success rate averaged across tasks. As shown in Table \ref{table:ablation}, all these designs are significant for the performance of \textit{DemoSpeedup}. 

\subsection{Visualization Analysis}
To delve deeper into what patterns the entropy captures, we visualize the entropy curve alongside snapshots from the corresponding demonstrations. As shown in Figure \ref{fig:entropy vis}, the entropy and our segment approach effectively distinguish precise skills from nonchalant movements. Most Motions that approaches an object (Mark 2) or move an object in the air (Mark 5) are recognized as impeccable part. For precision part, the entropy curve could recognize not only contact-rich skills, like picking the yogurt (Mark 1) and transferring the chopsticks (Mark 3), but also contact-free motions such as carefully withdrawing the gripper from the inserted plate to prevent knocking over the plate (Mark 6), or cautiously aligning the chopsticks with the narrow box gap (Mark 4). 
\section{Conclusion}
In this paper, we present \textit{DemoSpeedup}, a self-supervised method to accelerate visuomotor policy execution. \textit{DemoSpeedup} leverages the action entropy of the data estimated from a trained generative policy to guide the acceleration of demonstrations. A clustering-based scheme is proposed to segment the demonstrations into different precision levels according to the entropy. Then those segments are down-sampled at rates that increase with the entropy. Our experiments demonstrate the \textit{DemoSpeedup} can achieve remarkable speedup while maintaining the task performance across different imitation learning algorithms and robot platforms.

\textbf{Limitations.} There are several limitations of this work. First, though \textit{DemoSpeedup} could improve the success rate in some tasks, it occasionally causes minor performance drops, probably because of the dynamics mismatch between the original and accelerated demonstrations. 
Second, as a self-supervised approach, the \textit{DemoSpeedup} pipeline avoids the trouble of human supervision. However, due to the inherent variations in the execution speed of datasets collected by different human operators, the potential for acceleration also varies. As a result, the desired acceleration rate in \textit{DemoSpeedup} needs to be manually determined.  
Finally, this work doesn't consider the DP inference delay that has an influence on execution acceleration. This can be solved using distillation methods\citep{prasad2024consistency,lu2024manicm} or flow-based policies\citep{black2410pi0}.




\clearpage
\acknowledgments{We would like to give special thanks to Galaxea Inc. for hardware support and maintenance, Zhenyu Jiao, Ke Dong and Yixiu Li for their technical support on the controller, Ke Sheng and Zhenghao Qi for their advice on VR setup and imitation learning tuning on Galaxea R1. We also thank Kaizhe Hu for initial discussion on the project's direction  and Zhecheng Yuan for helping data collection. Tsinghua University Dushi Program supports this project.}


\bibliography{example}  
\newpage
\section*{Appendix}
\section{\textit{DemoSpeedup} pseudocode}
 
We provide the complete pseudocode of \textit{DemoSpeedup} in Algorithm \ref{alg:demospeedup}.

\begin{figure}[h]
\vspace{-20pt}
    \begin{algorithm}[H]
        \SetAlgoNoLine
        \textbf{Input}: $\mathcal{D} = \{\tau_i\}_{i=1}^B,  \tau_i=\{o_t,a_t\}_{t=1}^T$; {\color{olive} // original demonstrations}\\
        \textbf{Train $\pi_{proxy}$ on $\mathcal{D}$}; {\color{olive} // train proxy policy on original demonstrations with ACT or DP}\\
        
        {\color{olive} // action entropy estimation via proxy policy}\\
        \texttt{ def} \textit{get\_entropy}$(\pi_{proxy}, \tau_i)$: 

        \Indp \textbf{Initialize: }$t=1$,\quad$\mathcal{H}_{list}=[\,]$,\quad$S=\{\}, N$\\
        {\color{olive} // $\mathcal{H}_{list}$ is the entropy list, $S$ is the action sample set, $N$ is the number of samples}\\
        \While{$t < T$}{
            \Indp \For{$i \gets 1$ to $N$}{
            $\hat{A}_t^i \xleftarrow{} \pi_{proxy}(A_t|o_t,z_i)$;{\color{olive} // $A_t$ is the action chunk, $z_i$ is sampled latent variable for ACT and sampled noise for DP}\\
            Add action samples in $\hat{A}_t^i$ to $S$;            
            }
            Calculate $\mathcal{H}_t$ according to Equation\ref{equation1},\ref{equation2};\\
            Add $\mathcal{H}_t$ to $\mathcal{H}_{list}$;
            
    }
   
        \texttt{\textbf{return}} $\mathcal{H}_{list}$;\\
     \Indm{\color{olive} // accelerate demos with entropy}\\
    \texttt{ def} \textit{accelerate\_demos}$(\mathcal{H}_{list}, \mathcal{D})$: 

        \Indp \textbf{Preprocess: } $\mathcal{H}_{list}\xleftarrow{}Isolation\_Forest(\mathcal{H}_{list})$;\\
        \textbf{Label Precision: }
        $\{P,C\} \xleftarrow{} Hdbscan\_Cluster(\mathcal{H}_{list})$;\\
         \textbf{Initialize: } $\mathcal{D}_{speedup}=\{\tau^{speedup}_i\}_{i=1}^B$, $\tau^{speedup}_i=[\,]$,$K$ \\{\color{olive} // $K$ is the chunk size of accelerated policy}\\
        \For{$i \gets 1$ to $B$}{
           Sample $\tau_i$ from $\mathcal{D}$;
        \Indp \For{$t \gets 1$ to $T$}{
           Sample $\{o_t,a_{t:T}\}$ from $\tau_i$;
           $A_t^{speedup}\xleftarrow{}{piecewise\_downsample\_actions(a_{t:T},\{P,C\})}$;\\
           Add $\{o_t, A_t^{speedup}[:K]\}$ to $\tau^{speedup}_i$;
        }
        }
        \texttt{\textbf{return}} $\mathcal{D}_{speedup}$;\\

    \Indm{\color{olive} // down-sample actions in sub-trajectories with precision label guidance}\\
    \texttt{ def} \textit{piecewise\_downsample\_actions}$(a_{t:T}, \{P,C\})$: 

        \Indp \textbf{Initialize: } $r_{high},r_{low},indices=[\,]$; {\color{olive} //  $r_{high},r_{low}$ is high and low down-sample ratio}\\
        \For{$i \gets t$ to $T$}{
        \uIf{$[i:{i+r_{high}}]\subseteq C$}{
            $i \gets i + r_{high}$, Add $i$ to $indices$\;
        }
        \Else{
            $i \gets i + r_{low}$, Add $i$ to $indices$\;
        }
    
        }
        $A_t^{speedup}\xleftarrow{}a_{indices}$;\\
        \texttt{\textbf{return}} $A_t^{speedup}$;\\  
    \Indm\textbf{Train $\pi_{speedup}$ on $\mathcal{D}_{speedup}$ by directly imitating $\{o_t, A_t^{speedup}[:K]\}$};\\
    {\color{olive} // train accelerated policy on accelerated demonstrations with ACT or DP}\\
        \textbf{output}: $\pi_{speedup}$
        \caption{\textit{DemoSpeedup} for accelerating demonstrations to train visuomotor policy. }
        \label{alg:demospeedup}
    \end{algorithm}
\end{figure}

\section{Additional Comparisons}
\subsection{Comparison with other demonstration speedup methods}
\label{app: comparison with more}
We further compare the performance of \textit{DemoSpeedup} with other down-sample baselines by replacing the entropy guided piecewise acceleration with following methods:
\begin{itemize}[leftmargin=20pt]
    \item \textit{Contact Oracle}: We design a contact-based heuristic to segment low-precision and high-precision subtrajectories. The division rule is as follows: whenever a new object-pair contact or detachment occurs in the Manipulation scene (e.g., the gripper contacts with the object, or one object contacts with another object), a constant time before and after the moment when the contact state changes are labeled as precision. The rest of the time is labeled as casualness. Note that this method requires oracle information and precise 3D priors, which are difficult to obtain in real-world settings with only 2D camera inputs.
    \item \textit{AWE*}:  We adjust a dynamic programming method from AWE\citep{shi2023waypoint} to down-sample the data. AWE aims to promote success rate. It minimizes the trajectory length by $7\!\!\times\!\!-\!10\times$ given a threshold constraint of piece-wise linear approximation error of the joint-angle trajectories. AWE needs much longer time than demonstrations to reach waypoints. So we tune the threshold to reduce the trajectory length to roughly $2\times$ and use the same control frequency as demonstrations for acceleration. Besides, AWE only relies on joint-angle trajectories which can be noisy and task-irrelevant. We improve it by re-weighting the approximation error with entropy.
    \item \textit{Constant 2$\times$}: Directly down-sample the demonstrations at $2\times$ ratio.
    \item \textit{Constant 3$\times$}: Directly dow-sample the demonstrations at $3\times$ ratio.

\end{itemize}

\begin{table}[htbp]
\centering

\setlength{\tabcolsep}{3.5pt} 
\begin{tabularx}{\textwidth}{@{}l *{8}{>{\centering\arraybackslash}X}@{}} 
\toprule
\textbf{Method \& Algo} 
& \multicolumn{2}{c}{\textbf{Transfer Cube\&ACT}} 
& \multicolumn{2}{c}{\textbf{Insertion\&ACT}} 
& \multicolumn{2}{c}{\textbf{Transfer Cube\&DP}} 
& \multicolumn{2}{c}{\textbf{Insertion\&DP}} \\
\cmidrule(lr){2-3} \cmidrule(lr){4-5} \cmidrule(lr){6-7} \cmidrule(lr){8-9}
\addlinespace[-8pt]
& \makecell{success \\[-2.5pt] rate ($\uparrow$)} 
& \makecell{episode \\[-2.5pt] len ($\downarrow$)} 
& \makecell{success \\[-2.5pt] rate ($\uparrow$)} 
& \makecell{episode \\[-2.5pt] len ($\downarrow$)} 
& \makecell{success \\[-2.5pt] rate ($\uparrow$)} 
& \makecell{episode \\[-2.5pt] len ($\downarrow$)} 
& \makecell{success \\[-2.5pt] rate ($\uparrow$)} 
& \makecell{episode \\[-2.5pt] len ($\downarrow$)} \\
\addlinespace[-8pt]
\midrule
\textit{Origin}   & \singleddbf{40\%} & \singledd{321} & \singledd{11\%} & \singledd{435} & \singledd{47\%} & \singledd{289} & \singledd{12\%} & \singledd{329} \\
\textit{Contact Oracle}   & \singledd{37\%} & \singledd{140} & \singledd{15\%} & \singledd{142} & \singledd{37\%} & \singledd{124} & \singledd{11\%} & \singleddbf{127} \\
\textit{DemoSpeedup}   & \singleddbf{40\%} & \singleddbf{137} & \singleddbf{22\%} & \singleddbf{125} & \singleddbf{49\%} & \singleddbf{121} & \singleddbf{16\%} & \singledd{145} \\
\bottomrule
\vspace{0.4pt}
\end{tabularx}
\caption{\textbf{Comparison with \textit{Contact Oracle. }}We collect a new dataset including contact information using a trained checkpoint to conduct the experiment. \textit{DemoSpeedup} achieves a comparable success rate with \textit{Origin} while \textit{Contact Oracle} often performs worse than \textit{Origin}.} 
\label{tbl:compare-with-contact}
\end{table}
\vspace{-14pt}
\begin{table}[htbp]
\centering

\setlength{\tabcolsep}{3.5pt} 
\begin{tabularx}{\textwidth}{@{}l *{8}{>{\centering\arraybackslash}X}@{}} 
\toprule
\textbf{Method \& Algo} 
& \multicolumn{2}{c}{\textbf{Transfer Cube\&ACT}} 
& \multicolumn{2}{c}{\textbf{Insertion\&ACT}} 
& \multicolumn{2}{c}{\textbf{Transfer Cube\&DP}} 
& \multicolumn{2}{c}{\textbf{Insertion\&DP}} \\
\cmidrule(lr){2-3} \cmidrule(lr){4-5} \cmidrule(lr){6-7} \cmidrule(lr){8-9}
\addlinespace[-8pt]
& \makecell{success \\[-2.5pt] rate ($\uparrow$)} 
& \makecell{episode \\[-2.5pt] len ($\downarrow$)} 
& \makecell{success \\[-2.5pt] rate ($\uparrow$)} 
& \makecell{episode \\[-2.5pt] len ($\downarrow$)} 
& \makecell{success \\[-2.5pt] rate ($\uparrow$)} 
& \makecell{episode \\[-2.5pt] len ($\downarrow$)} 
& \makecell{success \\[-2.5pt] rate ($\uparrow$)} 
& \makecell{episode \\[-2.5pt] len ($\downarrow$)} \\
\addlinespace[-8pt]
\midrule
\textit{Origin}   & \singledd{72\%} & \singledd{291} & \singledd{21\%} & \singledd{452} & \singledd{66\%} & \singledd{281} & \singledd{16\%} & \singledd{431} \\
\textit{AWE*}   & \singledd{63\%} & \singledd{148} & \singledd{14\%} & \singledd{183} & \singledd{53\%} & \singledd{169} & \singledd{9\%} & \singledd{221} \\
\textit{Constant $2\times$}   & \singledd{80\%} & \singledd{167} & \singledd{27\%} & \singledd{242} & \singleddbf{75\%} & \singledd{152} & \singledd{20\%} & \singledd{247} \\
\textit{Constant $3\times$}   & \singledd{47\%} & \singledd{126} & \singledd{7\%} & \singledd{163} & \singledd{39\%} & \singledd{109} & \singledd{4\%} & \singleddbf{198} \\
\textit{DemoSpeedup}   & \singleddbf{81\%} & \singleddbf{121} & \singleddbf{30\%} & \singleddbf{151} & \singledd{74\%} & \singleddbf{107} & \singleddbf{29\%} & \singledd{218} \\
\bottomrule
\vspace{0.4pt}
\end{tabularx}
\caption{\textbf{Comparison with other baselines. }Compared to other down-sample methods, our approach achieves the best balance between success rate and speed. } 
\label{tbl:compare-with-other}
\end{table}
\vspace{-8pt}

Additionally, we utilize \textit{Origin} to refer policies trained on original demonstrations. Other factors including replicate-before-downsample and Geometrical consistency are the same to guarantee a fair comparison. We conduct experiments on Aloha with ACT and DP. To compare with \textit{Contact Oracle}, since the original datasets doesn't offer any privileged information, we recollect 50 new demonstrations including the contact information by rollouting a trained ACT. Then \textit{Contact Oracle}, \textit{Origin} and \textit{DemoSpeedup} are all trained on this dataset for comparison. To compare with other methods, we still use the original dataset.

The comparison results with \textit{Contact Oracle} are shown in Table \ref{tbl:compare-with-contact}. \textit{DemoSpeedup} achieves a better performance than \textit{Contact Oracle} while achieving similar speedup. \textit{Contact Oracles} often falls short of the  success rate of \textit{Origin}. We argue that this is because contact pattern can't account for all precision patterns and could only offer a rough estimation of precision. For example, in Insertion task, the contact pattern keeps the same after one block first contacts with the other block, thus failing to capture the contact-rich phase of one block being inserted to the other. Besides, the constant time period around the contact change moment fails to offer an accurate estimation of precision stage duration.

The comparison results with other methods are shown in Table \ref{tbl:compare-with-other}. \textit{DemoSpeedup} strikes the best balance between the success rate and speedup. Specifically, \textit{DemoSpeedup} achieves a similar performance with \textit{Constant $2\times$} and a similar speedup with \textit{Constant $3\times$}. Additionally, though AWE could promote success rate in its original setting, we find \textit{AWE*} even worse than \textit{Constant $2\times$} for acceleration. It is mostly because the approach is based on dynamic programming which doesn't consider the smoothness of selected actions. Therefore, the accelerated policy often produces jittery motions, which hurts the performance. This demonstrates the unique challenge of accelerating policy execution that is different from traditional down-sample settings.

\subsection{Comparison with traditional down-sample approaches}
Down-sampling the demonstrations has been a widely used practice in robot community. However, most down-sample approaches serve for improving the performance rather than accelerating policy execution. \textit{DemoSpeedup} differs from previous approaches in following two ways:
\begin{itemize}[leftmargin=20pt]
 \item \textbf{Execution frequency. }Our down-sample method decreases the demonstration frequency but doesn't decrease the policy execution frequency. For example, for a 50Hz recorded demonstration, traditional methods may down-sample it to 20Hz and deploy the trained policy at 20Hz. However, in our setting, we down-sample it at 20Hz but deploy the checkpoint at original 50Hz in order to accelerate execution.
 \item \textbf{Novel Challenge. }The main challenge in this work is to maintain the performance while speeding up the execution. Thus even a $< 5\%$ drop in success rate is intolerable. Besides, previous methods such as Keyframes\citep{shridhar2023perceiver} rely on close-loop control to reach down-sampled waypoints, so the speed is even slower than original demonstrations. On the contrary, the execution speed in this work is much faster than the demonstrations. Thus, the challenge of acceleration demands a much higher accuracy in recognizing precision patterns than traditional settings. Traditional heuristic methods perform poorly against this challenge, as shown in \ref{app: comparison with more}. \textit{DemoSpeedup} well mitigates this challenge using entropy and maintains the performance.
\end{itemize}

\section{Hyperparameters}

\subsection{High and low down-sample ratio}
\begin{table}[!htbp]

\vspace{-4pt}
\centering
\begin{tabular}{@{}llll@{}}
\toprule
\textbf{Task} & \textbf{$\{r_{low},r_{high}\}$} & \textbf{Task} & \textbf{$\{r_{low},r_{high}\}$} \\ 
\midrule
Transfer Cube          & \{2, 4\}                     & Open Trays             & \{2, 4\}               \\
Insertion         & \{2, 4\}                     & Flip Cutlery    & \{1, 3\}                \\
Sandwich Remove & \{2, 4\}                 & Cupboard Open & \{2, 4\}        \\
Move Plate   & \{2, 4\}                     & Pen in Cup    & \{2, 4\}                \\
Load Cups   & \{2, 4\}                       & Sort   & \{2, 4\}                \\
Put Cups   & \{2, 4\}                       & Kitchenware    & \{2, 4\}               \\
Saucepan to Hob   & \{2, 4\}                      & Bomb Deposal    & \{2, 3\}                \\
Drawer Close   & \{2, 4\}                      & Conveyer    & \{2, 4\}               \\
\bottomrule
\end{tabular}
\vspace{1pt}
\caption{\textbf{Hyperparameter of high and low down-sample rate for ACT.}}
\label{supp:down-sample rate act}
\end{table}
\vspace{-8pt}
The key hyperparameters in \textit{DemoSpeedup} is $r_{high}$ and $r_{low}$, the high and low down-sample ratio. They directly impact the performance and speedup of accelerated policy. We keep them the same in most tasks, proving the robustness and generalization of our approach. However, if some non-accelerated demonstrations are fast enough or the task requires extremely precision, we need to manually tune them via several trials. $r_{high}$ and $r_{low}$ are shown in Table \ref{supp:down-sample rate act} and \ref{supp:down-sample rate dp}. Empirically, selecting down-sample rate is relatively easy, as we only use integers as the down-sample ratio. One can train checkpoints for multiple ratios simultaneously and evaluate which works best.

\begin{table}[!tbp]

\vspace{-4pt}
\centering
\begin{tabular}{@{}llll@{}}
\toprule
\textbf{Task} & \textbf{$\{r_{low},r_{high}\}$} & \textbf{Task} & \textbf{$\{r_{low},r_{high}\}$} \\ 
\midrule
Transfer Cube          & \{2, 4\}                     & Open Trays             & \{2, 4\}               \\
Insertion         & \{2, 4\}                     & Flip Cutlery    & \{1, 3\}                \\
Sandwich Remove & \{2, 4\}                 & Cupboard Open & \{2, 4\}        \\
Move Plate   & \{2, 3\}                     & Pen in Cup    & \{2, 4\}                \\
Load Cups   & \{2, 4\}                       & Sort   & \{2, 4\}                \\
Put Cups   & \{2, 3\}                       & Kitchenware    & \{2, 3\}               \\
Saucepan to Hob   & \{2, 4\}                      & Bomb Deposal    & \{2, 3\}                \\
Drawer Close   & \{2, 4\}                      & Conveyer    & \{2, 4\}               \\
\bottomrule
\end{tabular}
\vspace{1pt}
\caption{\textbf{Hyperparameter of high and low down-sample rate for DP.}}
\label{supp:down-sample rate dp}
\end{table}

\subsection{ACT Hyperparameters}
We utilize the same hyperparameter configuration for ACT across all tasks, shown in Table \ref{table:hparam_act}. For chunk length, we utilize time duration to represent it, as different tasks in this work have different control frequencies. The specific chunk size is the chunk length multiplied by the control frequency. For ACT+\textit{DemoSpeedup}, we use half of ACT's chunk length to ensure geometrical consistency. This is based on the observation that most speedups in our experiments are around $2\times$.
\begin{table*}[tbh!]
\centering
\setlength{\tabcolsep}{32pt}
\begin{tabular}{lll}
\toprule
\textbf{Hyperparameter} & \textbf{ACT} & \textbf{ACT + \textit{DemoSpeedup}} \\ \midrule
learning rate & 1e-5 & 1e-5 \\
\# encoder layers & 4 & 4 \\
\# decoder layers & 7 & 7 \\
feedforward dimension & 3200 & 3200 \\
hidden dimension & 512 & 512 \\
\# heads & 8 & 8 \\
chunk length & 1s & 0.5s \\
beta & 10 & 10 \\
dropout & 0.1 & 0.1 \\

\bottomrule
\end{tabular}
\caption{\textbf{ Hyperparameters of ACT + \textit{DemoSpeedup} and ACT. }The only difference is the reduction in chunk size.}
\label{table:hparam_act}
\end{table*}

\begin{table*}[h]
\centering

\begin{tabular}{l|l|l|l}
\toprule
\textbf{Hyperparameter}        & \textbf{Aloha} & \textbf{Bigym} & \textbf{Real-world} \\
\midrule
observation horizon             & 1    & 1    & 1    \\
diffusion\_step\_embed\_dim    & 128    & 256    & 128    \\
down\_dims  & [512,1024,2048]   & [256,512,1024]   & [512,1024,2048]   \\
kernel\_size & 5 & 5 & 5 \\
n\_groups & 8 & 8 & 8 \\
vision\_model     & Resnet18   & MVT\citep{seo2023multi}   & Resnet18   \\
chunk size         & 48    & 16    & 24    \\
Lr            & 1e-4 & 1e-4 & 1e-4 \\
WDecay         & 1e-6 & 1e-6 & 1e-6 \\
scheduler   & DDIM  & DDIM  & DDIM  \\
$D$-Iters Train  & 100  & 100  & 100  \\
$D$-Iters Eval   & 10  & 10  & 10  \\

\bottomrule
\end{tabular}

\caption{
\textbf{Hyperparameters for Diffusion Policy}.
\label{tab:dp_transformer}
}
\vspace{-5mm}
\end{table*}

\subsection{Diffusion Policy Hyperparameters}
We utilize separate hyperparameter configurations across Aloha, Bigym and real world for best performance, shown in Table \ref{tab:dp_transformer}, \ref{tab:dp_demospeedup_hyperparam}. All the DP used in our experiments are CNN-based. 
\begin{table*}
\centering

\begin{tabular}{l|l|l|l}
\toprule
\textbf{Hyperparameter}        & \textbf{Aloha} & \textbf{Bigym} & \textbf{Real-world} \\
\midrule
observation horizon             & 1    & 1    & 1    \\
diffusion\_step\_embed\_dim    & 128    & 256    & 128    \\
down\_dims  & [512,1024,2048]   & [256,512,1024]   & [512,1024,2048]   \\
kernel\_size & 5 & 5 & 5 \\
n\_groups & 8 & 8 & 8 \\
vision\_model     & Resnet18   & MVT\citep{seo2023multi}   & Resnet18   \\
chunk size         & 24    & 8    & 12    \\
Lr            & 1e-4 & 1e-4 & 1e-4 \\
WDecay         & 1e-6 & 1e-6 & 1e-6 \\
scheduler   & DDIM  & DDIM  & DDIM  \\
$D$-Iters Train  & 100  & 100  & 100  \\
$D$-Iters Eval   & 10  & 10  & 10  \\

\bottomrule
\end{tabular}

\caption{
\textbf{Hyperparameters for DP+\textit{DemoSpeedup}.} The only difference with DP is the reduction in chunk size.
\label{tab:dp_demospeedup_hyperparam}
}
\vspace{-5mm}
\end{table*}
\end{document}